# Semantic Segmentation via Highly Fused Convolutional Network with Multiple Soft Cost Functions

Tao Yang, Yan Wu[*], Junqiao Zhao, Linting Guan

College of Electronics & Information Engineering, Tongji University,
201804, Shanghai, China

`yanwu@tongji.edu.cn`

## Abstract

Semantic image segmentation is one of the most challenged tasks in computer vision. In this paper, we propose a highly fused convolutional network, which consists of three parts: feature downsampling, combined feature upsampling and multiple predictions. We adopt a strategy of multiple steps of upsampling and combined feature maps in pooling layers with its corresponding unpooling layers. Then we bring out multiple pre-outputs, each pre-output is generated from an unpooling layer by one-step upsampling. Finally, we concatenate these pre-outputs to get the final output. As a result, our proposed network makes highly use of the feature information by fusing and reusing feature maps. In addition, when training our model, we add multiple soft cost functions on pre-outputs and final outputs. In this way, we can reduce the loss reduction when the loss is back propagated. We evaluate our model on three major segmentation datasets: CamVid, PASCAL VOC and ADE20K. We achieve a state-of-the-art performance on CamVid dataset, as well as considerable improvements on PASCAL VOC dataset and ADE20K dataset.

## 1. Introduction

In the past ten years, with large quantities of labeled datasets, deep neural networks have been widely used in processing of image, natural language and speech. Among these deep networks, convolutional neural network (CNN) plays a most important role. Convolutional neural network has shown its outstanding performance in many fields, a series of CNN-based networks and some useful independent modules have been brought forward too, such as dropout [1] and batch normalization [2]. Convolutional networks are now leading many computer vision tasks, including image classification [3, 4], object detection [5, 6, 7, 8] and semantic image segmentation [9, 10, 11]. Image semantic segmentation is also known as scene parsing, which aims to classify every pixel in the image. It is one of the most challenged and primary tasks in computer vision. Network models for scene parsing task are always based on reliable models for image classification, since segmentation datasets have fewer images than the large available classification datasets. The landmark fully convolutional network (FCN) [9] for semantic segmentation is based on VGG-net [12], which is trained on the famous ImageNet dataset [13]. A novel end-to-end segmentation learning method is introduced in FCNs. In detail, convolution layers with a kernel size of 1x1 take the place of fully connected layers, followed by unpooling layers to recover the spatial resolution of the feature maps. As a consequence, output maps can achieve the same resolution as the input image of the models. In order to reduce the noise in output maps, FCN introduces skip connections between pooling layers and unpooling layers. Since the proposal of FCN, modern works on segmentation are mostly based on it [14, 15].

In our previous work, a fully combined convolutional network (FCCN) is explored to improve the segmentation performance [16]. We adopt a layer-by-layer upsampling method. After each upsampling operation, we acquire an output with the double size of the input feature maps. We also combine the corresponding pooling and unpooling layers. Another important work in FCCN is the soft cost function used for training the model. Evaluated on CamVid dataset [17], FCCN achieves an improvement of 10 percentage points in performance compared to FCN8s, which is also compatible to state-of-the-art.

In this paper, we extend FCCN with a highly fused network. We independently recover each unpooling layer to a pre-output layer, which has a spatial resolution equal to the input image. Since there are five unpooling layers in FCCN, our network comes out with five pre-output layers. Then a concatenation operation on these five pre-output layers is added, followed by a convolution layer to acquire the final output. Our fused network reuses the feature information more completely. In FCCN, the loss value is highly reduced after transferred back across the network, which results in the inefficient updating of parameters in lower layers. In order to resolve this problem, our proposed network is trained with multiple cost functions. We compute cost functions in both pre-output and final output layers, and the final cost function is a sum of these cost functions, with certain weights applied. With the multiple cost functions added on our fused network, feature information can be reused and parameters are updated more efficiently. Our network is evaluated on three datasets, CamVid [17], PASCAL VOC and ADE20K [18]. We achieve a considerable improvement in all these datasets, and even a state-of-the-art performance in CamVid dataset.

## 2. Related Work

Deep learning methods have become a ubiquitous technique in image segmentation. Compared to traditional computer vision methods, the key advantage of deep neural networks is the high capability of learning representation feature information with large pixel labeled datasets [19]. What is more, with end-to-end style of learning, deep networks require less domain expertise, effort than hand-crafted feature extraction methods. In recent years, image segmentation has been drawn great attention due to its huge potential in autonomous driving, virtual reality and robotic tasks [19]. As a result, semantic segmentation has achieved great progress. State-of-the-art methods on image segmentation are highly relied on CNN models trained on large labeled datasets.

In modern deep models, FCN [9] is seemed as the basic method of most later proposed models for image segmentation. Actually, there had been some preliminary attempts applying convolutional networks to pixel-wise classification before FCN was proposed [20]. However, FCN firstly took the advantages of existing CNN models as a powerful method to learn hierarchies of segmentation features. In fact, FCN can be considered as a milestone of segmentation task, not only because of its great improvements in performance, but also because it showed that CNN can efficiently learn how to make dense class predictions for semantic segmentation. After FCN, recent proposed models are mainly designed by (1) bringing out novel decoder structure of the networks [10, 11]; (2) adopting more efficient basic classification models [21, 22]; (3) adding integrating context knowledge with some independent modules [23, 24].

Though FCN-based decoder structures were considered most popular and successful in segmentation, there were also some other remarkable structures. SegNet [10] was a typical case of using an alternative decoder variant, in which an encoder-decoder convolution path was proposed. Method in [10] generated an index table to record the max value index in max pooling, and applied it to recover the feature maps in upsampling part. Noh et al. presented another deconvolution network [11] with a similar decoder path as SegNet, but they adopted deconvolution modules to implement upsampling operations. Ronneberger et al. added 2x2 up-convolution layer, with a concatenation with corresponding pooling layer in U-Net [25]. FCCN [16] could also be regarded as an alternative decoder structure.

Modern segmentation models are almost improved on powerful basic models, for example, FCN was a VGG-based segmentation model. Apart from VGG net, GoogleNet [4], ResNets [26] and DenseNets [27] showed their powerful capacities in image classification task too. As a result, these models were used as basic models to modern segmentation task. Lin et al proposed multi-path RefineNet [21], based on ResNets, and achieved a great success in major segmentation datasets. Jegou et al. [22] extended DenseNets to segmentation task, achieving state-of-the-art performance in urban scene benchmark dataset.

Though CNN has achieved great success in multiple vision tasks, features by CNN are limited for semantic segmentation since it requires both global information and local information. Therefore, some extra modules were introduced to deep convolutional networks to integrate context knowledge. Conditional Random Field (CRF) [28] was considered the most powerful traditional pixel classify methods, Chen et al. [14] added CRF to improve the boundaries of the segmentation result. Yu and Koltun [23] introduced a novel dilated convolution module to help expand the receptive field without losing resolution, with which multi-scale contextual information could be systematically aggregated. Another impressive structure was recurrent neural network (RNN), Visin et al. [24] proposed a ReSeg segmentation network that based on RNN and achieved appreciable improvements.

## 3. Highly Fused Convolutional Network

We have achieved considerable improvements by transforming FCN into fully combined network, FCCN, in [16]. FCCN adopted a structure of five unpooling layers, each unpooling layer upsampled the feature maps to a doubled resolution. Then a combination layer was added after each unpooling layer, which combined the unpooling layer and its corresponding pooling layer. With these combination layers, feature information acquired by lower layers could be reused and extended. After the combination layer, there was an extra convolution layer to extract further features and also decrease the channels of feature maps.

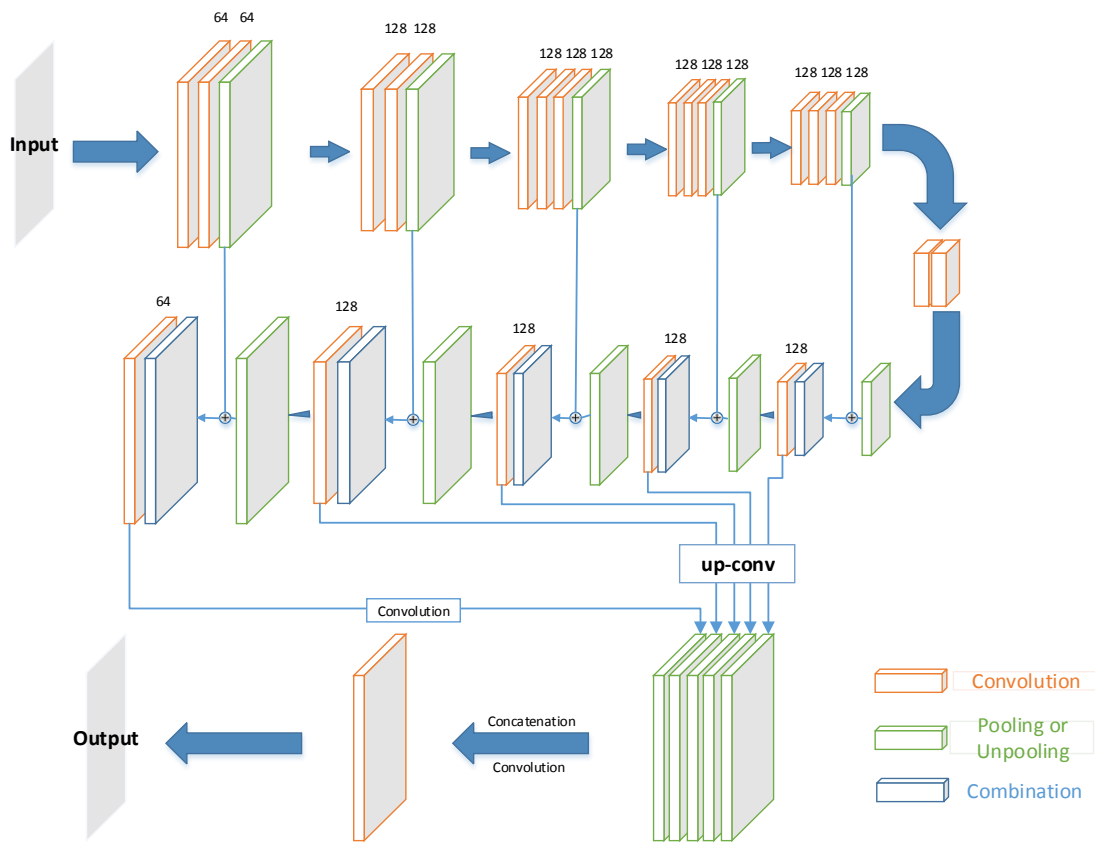

Fig. 1. Our highly fused network for segmentation task. The model can be divided into three parts: 1) the downsampling part helps to extract feature information, 2) the upsampling part recovers the spatial resolution of feature maps, 3) the final multi prediction part acquires efficient segmentation results.

Though the combined structure has achieved great improvements in segmentation performance, yet it cannot exhaustively use the extracted feature information. We have thought a lot over FCCN to come up with a further structured and efficient network, fortunately, we made it. We present our highly fused network in Fig. 1. As we can see, there are three major parts in our model: the downsampling part for extracting feature information, the upsampling part for recovering the spatial resolution, and the final multi prediction part for receiving efficient segmentation results.

The first downsampling part reserve the basic powerful classification model, with few modifications in filters channels. Just like FCN does, fully connected layers are transformed into convolution layers in basic VGG-16 model. In the upsampling part, we followed the combined structure in FCCN. As is showed in Fig. 1, each pooling layer is summed with its corresponding unpooling layer, feature maps in both are in the same spatial resolution. Fig. 2 is another view of the total combination path, which is a detailed representation for the combination operation. Then we also add a convolution layer for further non-linear feature extraction.

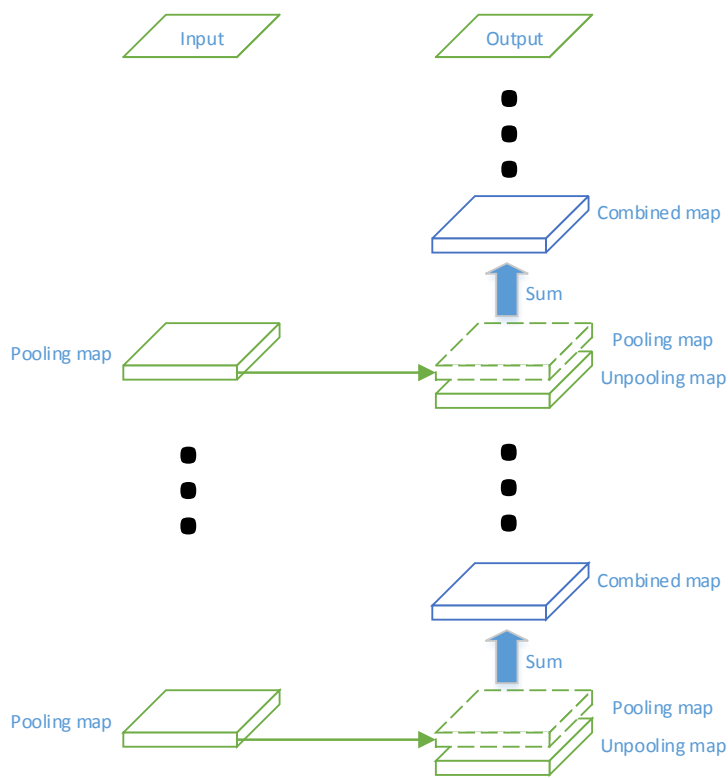

Fig. 2. Top-down sampling and down-top upsampling process, a combined feature map is the sum result of a pair of pooling feature map and its corresponding unpooling feature map. There are totally five pooling and unpooling layers.

We add an extra prediction part after the final upsampling layer. There are five unpooling blocks in the upsampling part, which reserve important feature information of different scales. Therefore, we add an up-conv block after each unpooling block, Fig. 3 shows what inside up-conv block. There is a one-step upsampling operation to recover the feature maps resolution, followed by convolution and nonlinear function RELU to generate the pre-output layer. Take a notice that the last unpooling block is already the same resolution as input image, thus we just need to convolve it to get the pre-output layer, without any upsampling operation needed. After these extra resolution recovering, we can achieve five pre-output layers, each with a channel of segmentation class number. We then concatenate these pre-output layers and convolve it to the final output layer.

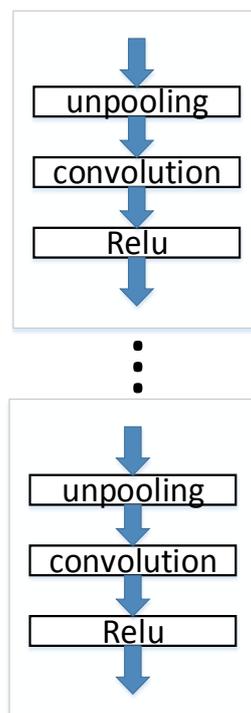

Fig. 3. Operations inside an up-conv block.

With the combination layers and concatenation of pre-output layers, our proposed model comes out as a highly fused convolutional network. The final prediction part makes use of multiple scale features in lower layers, which is of great importance in the final pixel-wise prediction. However, the gradients can be decreased rapidly due to transferring loss during back propagation. As a result, multiple cost functions are introduced and are explained in the next section.

## 4. Multiple Soft Cost Functions

Recent works on semantic segmentation mainly focus on modifying model structures, only a few researchers try to improve segmentation performance in modifying cost function when training their models. We find that small target areas are learned coarsely if no more attentions are paid on them. We should also be aware that in deep network models, gradients are always reduced or even frittered away in lower layers when transferred in back propagation. Thus, we try to do some work on cost function for more efficiently learning the network.

In FCCN, we have proposed a soft cost function for training a segmentation network [16]. Soft cost function adds a weight on each semantic area, while always keeps the weight background to be one. We dynamically calculate the weight of target semantic area according to its proportion in whole image. Thus, real weights of different semantic areas in one image can be calculated by:

$$w_{i,j} = \begin{cases} max\left(\frac{cb+\sum_{n=1}^{N} ct_n}{ct_n}, 2\right) & pixel(i, j) \in target\ class\ n \\ 1 & pixel(i, j) \in background \end{cases} \quad (1)$$

where $ct_n$ means pixel number of the n-th target class, $cb$ is the pixel number of background, and N is the number of classes. We set the weights on target semantic class always not smaller than 2, which promise that the target class acquire more loss in iterate training process. With the calculated weights, predicted value $H(i,j)$ and label value $Y(i,j)$ of pixel in position $(i,j)$, the final soft cost function is calculated as:

$$L = \sum_{i=1}^{m} \sum_{j=1}^{n} \frac{1}{2} w_{i,j} (H(i,j) - Y(i,j))^2 \quad (2)$$

Though network training focuses more on target learning with soft cost function, it is still unavoidable that the loss value will fritter away in back propagation. This is a vital hinder for learning parameters in lower layers efficiently. In order to decrease the loss reduction, we use multiple soft cost functions. We calculate a cost function on each pre-output layer and also the final output layer, thus there will be six cost functions in our training model. As we can know, cost functions in pre-output layers differ from the output layer, even in different pre-output layers the cost functions should be differed too, since they are recovered from feature maps of different spatial resolutions. As a result, we add a proportion $\lambda$ for each cost function. We should be aware that the final output layer is a concatenation result of pre-outputs, which means the final output layer contains more feature information than pre-outputs. Thus, we assign a constant proportion 1 to cost function $L_{fo}$ calculated on output layer, and proportions of other cost functions $L_{po_i}$, calculated on pre-output layers, are always not larger than 1. We get a final cost function like:

$$Loss = L_{fo} + \sum_{i=1}^{5} \lambda_i L_{po_i} \quad (3)$$

It is a challenged work for us to find out what proportions should be adopted, since we hardly had any prior idea about which pre-output should possess a larger proportion for training. We begin with a basic training with only one cost function on the final output layer, in other words, we set all proportions $\lambda_i$ to zero. Then we train our model with many groups of proportions, with each $\lambda_i$ is set from 1 to 0.1. In section 5.1 we present the detail result of different groups of proportions.

# 5. Experiments

In this section, we will show the performance that our model achieved on universal datasets CamVid, PASCAL VOC and ADE20K. We firstly evaluate our fused models and cost function proportions on CamVid dataset, and then extend it to the other two datasets. We implement our experiments on deep learning framework MatConvNet [29], an integrated matlab convolutional network toolbox. Respect to hardware, we use two GeForce GTX TITAN X and one GeForce GTX TITAN XP Pascal graphic cards. With cuda toolkit installed, MatConvNet can accelerate training process with graphic cards.

**Implementation details:** We firstly create a file of "*imdb.mat*" for each dataset, which integrate the necessary information of the dataset, such as images and labels path, images size and so on. Our models can easily acquire images for training, validation and testing by this file. When training our model, we choose a batch size of 10 images, with a learning rate of 1e-4, and come to an end after 50 epochs. The measure scores are evaluated by mean Intersection over Union (IoU) between ground truth and predicted semantic segmentation result:

$$\text{IoU} = \frac{true\ pos}{true\ pos + false\ pos + true\ neg} \quad (4)$$

## 5.1 CamVid

CamVid is the first collection of videos with object class semantic labels, complete with metadata [17]. It is acquired from 4 high quality videos captured at 30Hz, while the semantically labeled 11-class ground truth is provided at 1Hz. The dataset holds 367 frames for training, 101 frames for validation and 233 frames for test. Images in CamVid have a resolution of 960x720, in order to increase the training efficiency, we operate on a half-original resolution of 480x360.

Table 1. A comparison of performance on diverse cost function proportions with baseline FCN and FCCN

| Method | Proportion group | Mean IoU (%) | Mean pix.acc (%) | Pixel acc (%) |
| --- | --- | --- | --- | --- |
| FCNs[9] | - | 57.0 | 88.0 | - |
| FCCN[16] | - | 65.79 | 88.74 | 88.26 |
| Model1 | 0, 0, 0, 0, 0 | 68.16 | 89.31 | 88.05 |
| Model2 | 1, 1, 1, 1, 1 | 67.42 | 88.87 | 88.29 |
| Model3 | 0.5, 0.5, 0.5, 0.5, 0.5 | 67.95 | 88.97 | 88.58 |
| Model4 | 0.2, 0.2, 0.2, 0.2, 0.2 | 68.23 | 89.53 | 89.15 |
| Model5 | 0.1, 0.1, 0.1, 0.1, 0.1 | 68.02 | 89.25 | 88.70 |
| Model6 | 1, 0, 0, 0, 0 | 68.57 | 90.43 | 89.27 |
| Model7 | 0.5, 0.5, 0, 0, 0 | **69.94** | **92.61** | **89.79** |
| Model8 | 0.33, 0.33, 0.33, 0, 0 | 69.05 | 91.08 | 89.63 |
| Model9 | 0.25, 0.25, 0.25, 0.25, 0 | 68.46 | 89.94 | 89.16 |

We tried many groups of multiple cost functions proportions, Table 1 shows the test results of different proportions. We basically train our structured model with only one cost function on final outputs, which achieve a performance of 68.16% in mean IoU, an impressive increasing compared to our baseline FCN and FCCN. Then we set a constant proportion 1 to all cost functions, but we acquire a performance reduction with this group. We find that cost calculated by pre-output layers is much larger than cost on final output, while pre-output layers are more noisy than final output. Therefore, our model will learn more noise when training with the constant proportion group, and, as a result, decrease segmentation performance. Respect to this, we reduce the cost function proportions. From Table 1, we can see that the measured mean IoU increase as the proportion decrease at beginning, then reach the summit at model4, and start decreasing after then.

Nevertheless, it is an interesting discovery that the summation of proportions in best model is equal to cost proportion on final output. According to this discovery, we further train our network with proportion groups represented from model6 to model9. As showed in last four rows in Table 1, appreciable improvements are achieved in this way. We find that pre-outputs acquired by higher unpooling blocks are similar, which results to feature redundant. Therefore, we finally achieve a best performance on model7, which holds a proportion group of [0.5, 0.5, 0, 0, 0].

**Table 2.** Quantitative results of semantic segmentation on CamVid Dataset(%)

| Method | Building | Tree | Sky | Car | Sign | Road | Pedestrian | Fence | Pole | Sidewalk | Bicyclist | mIoU |
|---|---|---|---|---|---|---|---|---|---|---|---|---|
| SegNet [10] | 81.3 | 72 | **93** | 81.3 | 14.8 | 93.3 | 62.4 | 31.5 | 36.3 | 73.7 | 42.6 | 47.7 |
| Tripathi [30] | 74.2 | 67.9 | 91 | 66.5 | 23.6 | 90.7 | 26.2 | 28.5 | 16.3 | 71.9 | 28.2 | 53.2 |
| FCN8s[9] | 77.7 | 71.0 | 88.7 | 76.1 | 32.7 | 91.2 | 41.7 | 24.4 | 19.9 | 72.7 | 31.0 | 57.0 |
| DilatedNet [23] | 82.6 | 76.2 | 89.9 | 84 | 46.9 | 92.2 | 56.3 | 35.8 | 23.4 | 75.3 | 55.5 | 65.3 |
| FCCN [16] | 79.7 | 77.2 | 85.7 | 86.1 | 45.3 | 94.9 | 45.8 | 69.0 | 25.2 | 86.2 | 52.9 | 65.79 |
| Kundu [31] | 84 | 77.2 | 91.3 | 85.6 | 49.9 | 92.5 | 59.1 | 37.6 | 16.9 | 76 | 57.2 | 66.1 |
| FC-DenseNet [22] | 83.0 | 77.3 | 93.0 | 77.3 | 43.9 | 94.5 | **59.6** | 37.1 | 37.8 | 82.2 | 50.5 | 66.9 |
| Playing for data [32] | 84.4 | 77.5 | 91.1 | 84.9 | 51.3 | 94.5 | 59 | 44.9 | 29.5 | 82 | **58.4** | 68.9 |
| VPN-Flow[33] | - | - | - | - | - | - | - | - | - | - | - | 69.5 |
| Ours | **85.6** | **81.2** | **94.3** | **88.5** | **52.4** | **95.3** | 52.7 | **45.2** | **42.1** | **88.5** | 57.3 | **69.94** |

Table2 provides quantitative segmentation results on CamVid dataset. We outperform the state-of-the-art, achieved by VPN-Flow, by 0.44 percentage points. Respect to single category segmentation, ignore the VPN-Flow whose category segmentation is not published, we achieve the best performance on all categories except pedestrian and bicyclist. There always exist intersections between pedestrains and bicyclists, which make them hard to learn. In Fig 4, we present visual improvements of image segmentation on CamVid, from FCNs, FCCN to our proposed model.

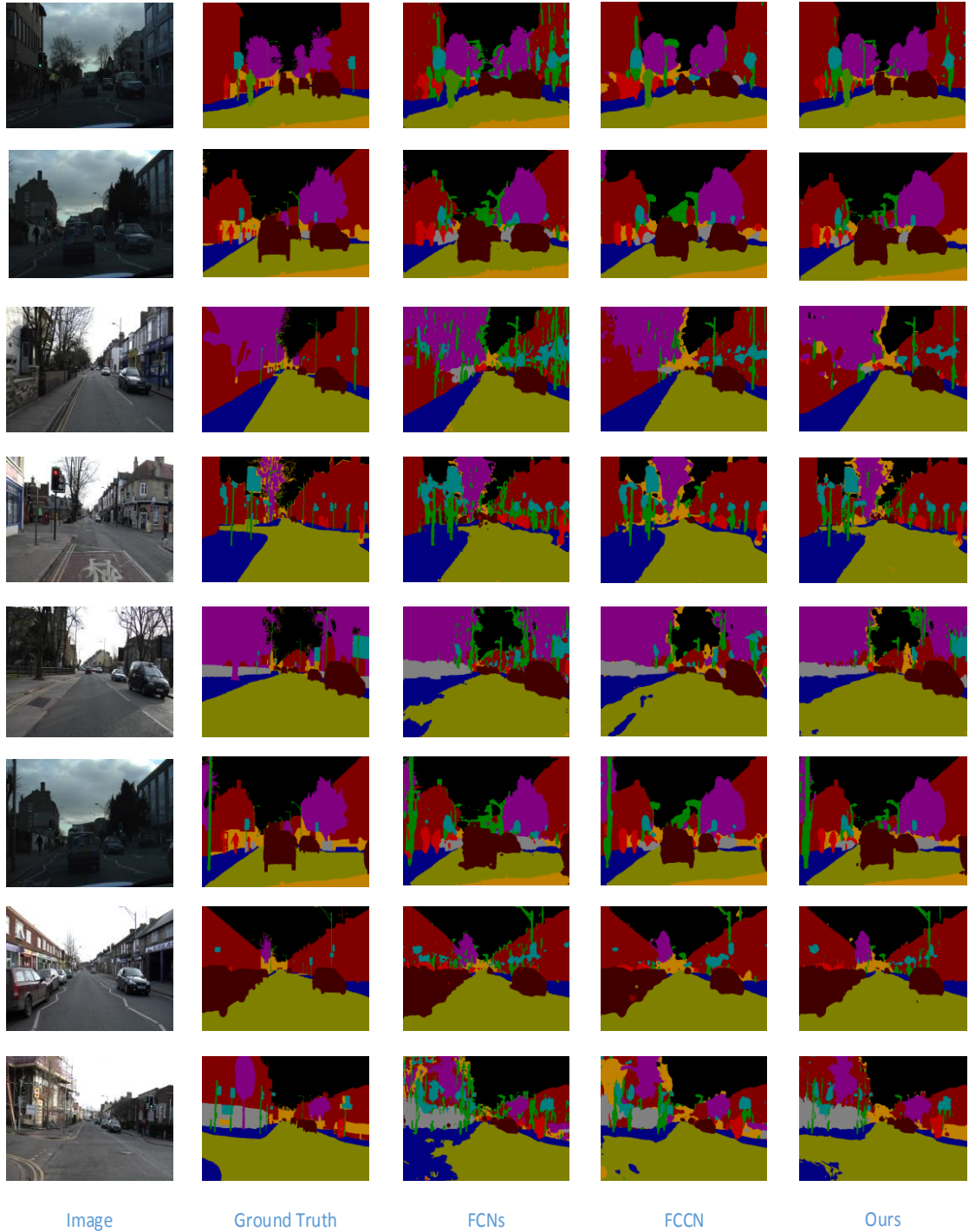

Fig. 4. Some visual segmentation examples on CamVid dataset. Compared to baseline FCNs and FCCN, our proposed method achieves more precise segmentation results.

## 5.2 ADE20K

ADE20K dataset contains more than 20K scene-centric images exhaustively annotated with objects and object parts. The benchmark is divided into 20K images for training, 2K images for validation, and another batch of held-out images for testing. There are totally 150 semantic categories included for evaluation, such as sky, road, grass, and discrete objects like person, car, and bed. Though ADE20K is a new scene parsing dataset, it is the most challenging one because of its complex scene classes. The data has been used in the Scene Parsing Challenge 2016 held jointly with ILSVRC'16.

**Table 3.** Our result on the validation set of MIT ADE20K with comparison with others.

| Method | Mean IoU (%) | Mean pix.acc (%) | Pixel acc (%) |
| --- | --- | --- | --- |
| FCNs [9] | 29.39 | 40.32 | 71.32 |
| SegNet [10] | 21.64 | 31.14 | 71.00 |
| DilatedNet [23] | 32.31 | 44.59 | 73.55 |
| Cascade-DilatedNet [18] | 34.90 | 45.38 | 74.52 |
| FCCN [16] | 36.43 | 49.76 | 76.31 |
| PSPNet [34] | 44.94 | - | 81.69 |
| Ours | 44.23 | 48.13 | 79.98 |

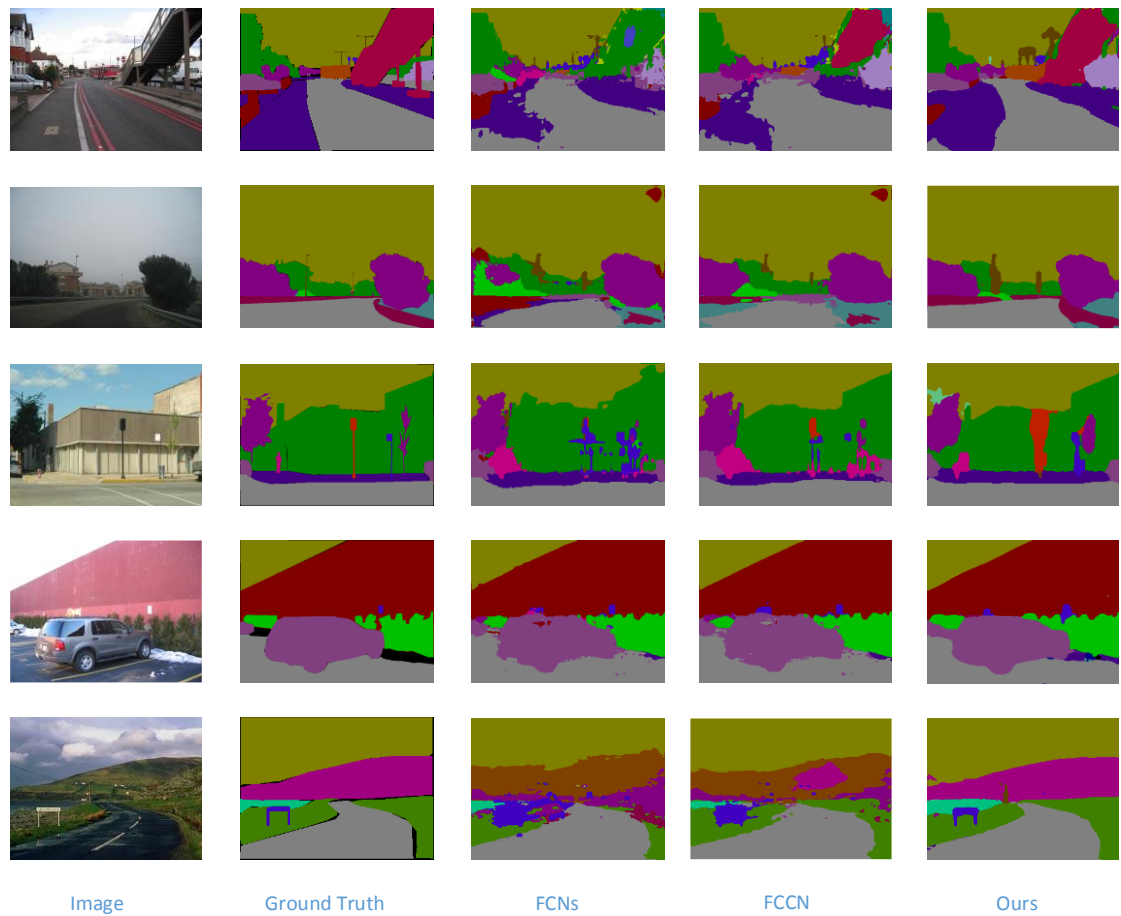

Image　　Ground Truth　　FCNs　　FCCN　　Ours

Fig. 5. Visual results on ADE20K dataset. We achieve more precise segmentation results than FCNs and FCCN.

We extend our work on this more challenged dataset. As shown in Table 3, compared to the state-of-the-art result, 44.94% in mean IoU, acquired by PSPNet, weachieve a comparable performance 44.23% on ADE20K dataset. While respect to our baseline FCCN, or even more basic FCNs, there comes out a giant increasing in performance. We increase the mean IoU by about 8 percentage points. In Fig. 5, we show some example predictions on the ADE20K dataset. We can find that we achieve more obvious segmentation results for target objects than FCNs and FCCN. Because of the large number of categories, segmentation in ADE20K seems still quite coarse. We shall overcome this shortcoming by using more powerful basic models in future work, such as ResNet.

## 5.3 PASCAL VOC

As we have achieved considerable improvement in CamVid dataset and ADE20K dataset, we further train our fused network on PASCAL VOC 2012 segmentation dataset. This dataset has a split of 10582, 1449 and 1456 images for training, validation and testing, which involves 20 object categories. PASCAL VOC challenge has held for 8 years, from 2005 to 2012, and has been one of the most momentous segmentation challenges.

There has been a lot of outstanding work on PASCAL VOC after FCN proposed. However, a very important point is that different basic models differ in classification ability, such as VGG net and ResNet. As a result, segmentation performance based on them also differs. In this paper, our work is based on VGG net, and Table 4 shows some of the best work on VGG based segmentation network. As we can see from Table 4, we achieve the best performance of 75.0 percentage in mean IoU. In most single category segmentation results, we get the highest score. Fig 6 shows some visual examples on PASCAL VOC dataset. Significant improvements in edge segmentation can be seen from Fig 6.

**Table 4.** Segmentation performance of VGG-based models on PASCAL VOC 2012(%)

| Method | aeroplane | bicycle | bird | boat | bottle | bus | car | cat | chair | cow | diningtable | dog | horse | motorbike | person | pottedplant | sheep | sofa | train | tvmonitor | mean IoU |
|---|---|---|---|---|---|---|---|---|---|---|---|---|---|---|---|---|---|---|---|---|---|
| FCNs [9] | 76.8 | 34.2 | 68.9 | 49.4 | 60.3 | 75.3 | 74.7 | 77.6 | 21.4 | 62.5 | 46.8 | 71.8 | 63.9 | 76.5 | 73.9 | 45.2 | 72.4 | 37.4 | 70.9 | 55.1 | 62.2 |
| FCCN [16] | 83.1 | 36.5 | 76.4 | 52.3 | 65.4 | 81.2 | 78.5 | 82.3 | 26.1 | 65.3 | 55.4 | 79.2 | 74.1 | 80.0 | 74.2 | 54.7 | 75.8 | 48.5 | 72.5 | 61.8 | 69.4 |
| Zoomout [15] | 85.6 | 37.3 | **83.2** | 62.5 | 66.0 | 85.1 | 80.7 | 84.9 | 27.2 | 73.2 | 57.5 | 78.1 | 79.2 | 81.1 | 77.1 | 53.6 | 74.0 | 49.2 | 71.7 | 63.3 | 69.9 |
| Piecewise [35] | 87.5 | 37.7 | 75.8 | 57.4 | **72.3** | 88.4 | 82.6 | 80.0 | 33.4 | 71.5 | 55.0 | 79.3 | 78.4 | 81.3 | 82.7 | 56.1 | 79.8 | 48.6 | 77.1 | 66.3 | 70.7 |
| DeepLab [14] | 84.4 | 54.5 | 81.5 | 63.6 | 65.9 | 85.1 | 79.1 | 83.4 | 30.7 | 74.1 | 59.8 | 79.0 | 76.1 | 83.2 | 80.8 | 59.7 | 82.2 | 50.4 | 73.1 | 63.7 | 71.6 |
| DeconvNet [11] | **89.9** | 39.3 | 79.7 | 63.9 | 68.2 | 87.4 | 81.2 | 86.1 | 28.5 | 77.0 | 62.0 | 79.0 | **80.3** | 83.6 | 80.2 | 58.8 | **83.4** | 54.3 | 80.7 | 65.0 | 72.5 |
| DPN [36] | 87.7 | 59.4 | 78.4 | 64.9 | 70.3 | 89.3 | 83.5 | 86.1 | 31.7 | **79.9** | 62.6 | 81.9 | 80.0 | 83.5 | 82.3 | 60.5 | 83.2 | 53.4 | 77.9 | 65.0 | 74.1 |
| ours | 88.5 | **60.3** | 82.4 | **65.7** | 70.5 | **89.8** | 84.2 | **87.3** | 33.9 | 78.4 | **64.1** | **82.3** | 79.8 | **83.7** | **83.5** | **62.4** | 81.7 | **55.6** | **81.4** | **67.2** | **75.0** |

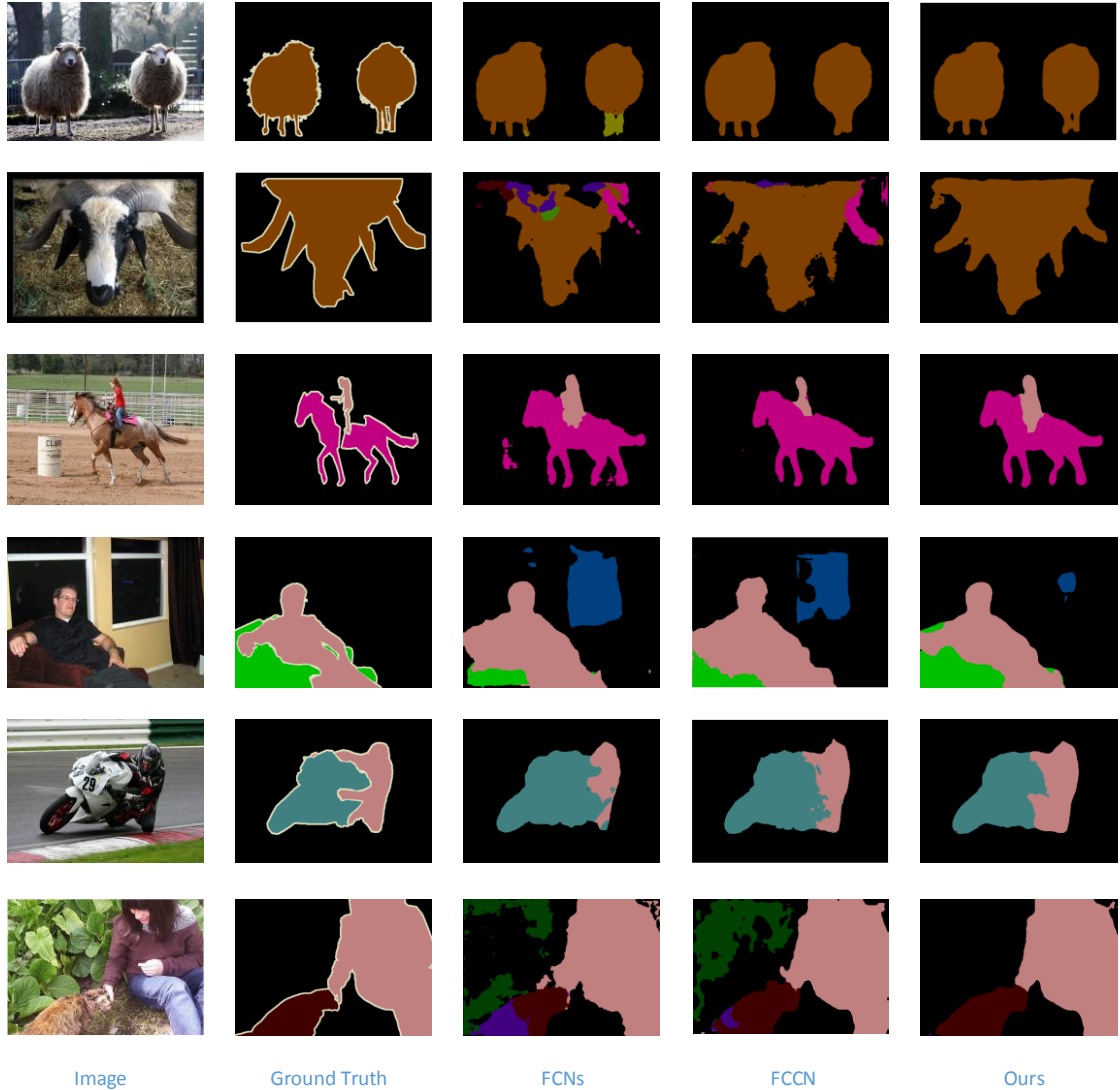

Fig. 6. Visual segmentation examples on PASCAL VOC 2012 dataset. Our proposed method achieves more precise results in edge segmentation.

## 6. Conclusion

In this paper, we proposed a highly fused segmentation network and validate its performance on three major semantic segmentation datasets. We adopt a downsampling and upsampling structure, with combinations between corresponding pooling and unpooling layers. We also add a prediction part in the end, which contains five pre-output layers and a final concatenated output. Our proposed fused network makes further use of feature information in low layers. Multiple soft cost functions are used to train our model, and considerable improvements are achieved in this way. In the future, we shall try more powerful basic classification model to increase the segmentation results. For example, we can adopt ResNet based segmentation network and apply our proposed multiple cost functions to train the model. We will also extend our work to some other fields in future work, such as apply it to lane detection. We believe that it would be a valuable and meaningful work.